%% file: acl2020.tex
\documentclass[11pt,a4paper]{article}
\usepackage[hyperref]{acl2020}
\usepackage{times}
\usepackage{latexsym}

\usepackage{graphicx}
\usepackage{microtype}
\usepackage{adjustbox}
\usepackage{booktabs}
\usepackage{makecell}
\usepackage{amsfonts}
\usepackage{amsmath}
\usepackage{url}
\usepackage{multicol}
\usepackage{multirow}
\usepackage{algorithm}
\usepackage{algorithmicx}
\usepackage{algpseudocode}

\aclfinalcopy 

\title{Unsupervised Dual Paraphrasing for Two-stage Semantic Parsing}

\author{Ruisheng Cao\qquad Su Zhu\qquad Chenyu Yang\\
\textbf{Chen Liu\qquad Rao Ma\qquad Yanbin Zhao\qquad Lu Chen\qquad Kai Yu}\thanks{\ \ The corresponding author is Kai Yu.}\\
  MoE Key Lab of Artificial Intelligence\\
  SpeechLab, Department of Computer Science and Engineering\\
  AI Institute, Shanghai Jiao Tong University, China\\
  {\tt \{211314,paul2204,yangcy,chris-chen,rm1031,zhaoyb\}@sjtu.edu.cn}\\
  {\tt \{chenlusz,kai.yu\}@sjtu.edu.cn}\\}
\date{}

\begin{document}
\maketitle

\input{abstract.tex}
\input{introduction.tex}

\input{training.tex}
\input{details.tex}
\input{experiment.tex}
\input{related.tex}
\input{conclusion.tex}

\section*{Acknowledgments}
We thank the anonymous reviewers for their thoughtful comments. 
This work has been supported by the National Key Research and Development Program of China (Grant No. 2017YFB1002102) and Shanghai Jiao Tong University Scientific and Technological Innovation Funds (YG2020YQ01).

\bibliography{acl2020}
\bibliographystyle{acl_natbib}

\clearpage
\appendix
\input{appendix.tex}

\end{document}

%% file: abstract.tex
\begin{abstract}
One daunting problem for semantic parsing is the scarcity of annotation. Aiming to reduce nontrivial human labor, we propose a two-stage semantic parsing framework, where the first stage utilizes an unsupervised paraphrase model to convert an unlabeled natural language utterance into the canonical utterance. The downstream naive semantic parser accepts the intermediate output and returns the target logical form. Furthermore, the entire training process is split into two phases: pre-training and cycle learning. Three tailored self-supervised tasks are introduced throughout training to activate the unsupervised paraphrase model. Experimental results on benchmarks \textsc{Overnight} and \textsc{GeoGranno} demonstrate that our framework is effective and compatible with supervised training.
\end{abstract}

%% file: introduction.tex
\section{Introduction}
\label{sec:intro}

\begin{figure}
    \centering
    \includegraphics[width=\columnwidth]{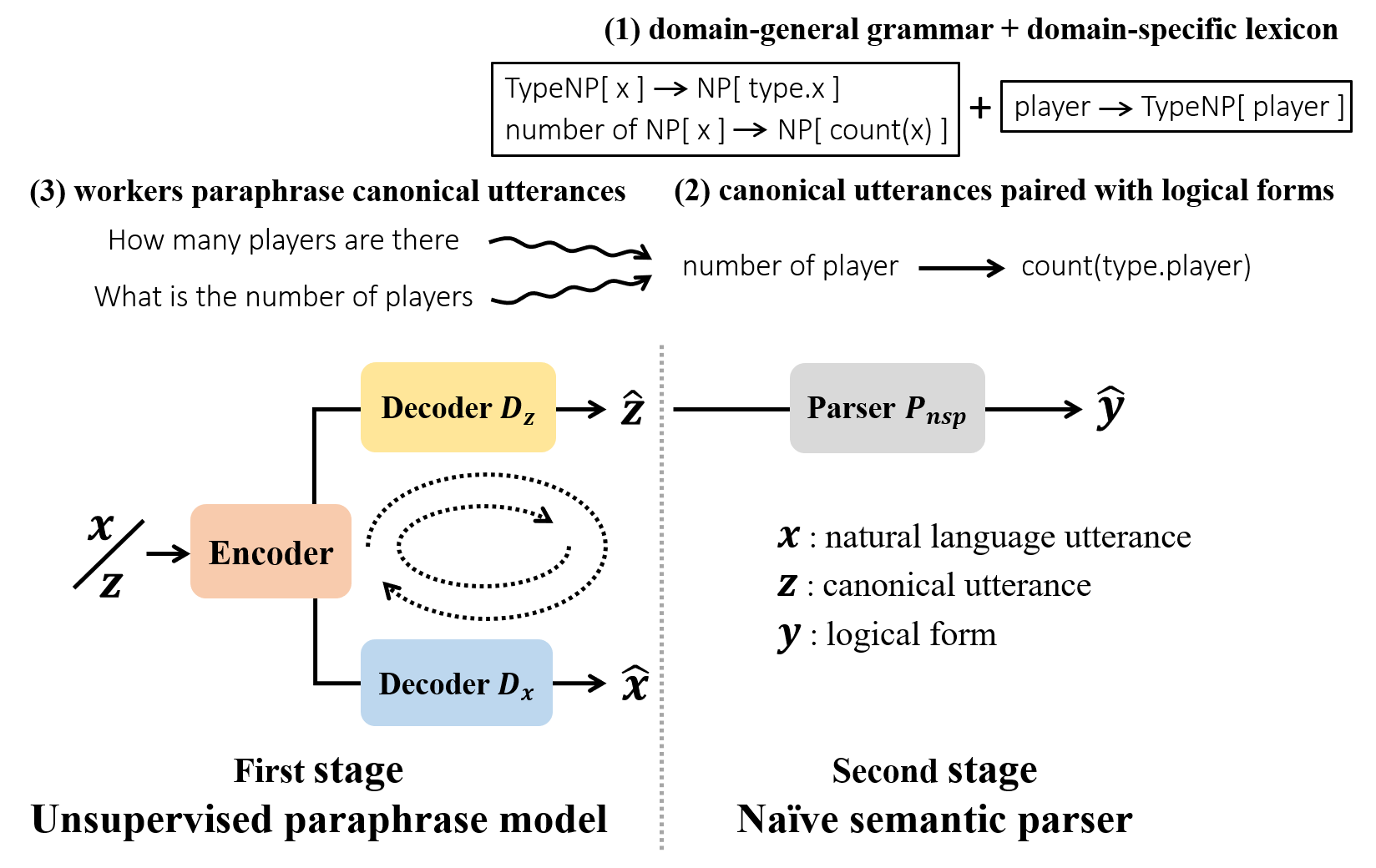}
    \caption{Two-stage semantic parsing framework, which is composed of an \emph{unsupervised} paraphrase model and a \emph{naive} neural semantic parser.}
    \label{fig:two-stage}
\end{figure}

Semantic parsing is the task of converting natural language utterances into structured meaning representations, typically logical forms~\citep{zelle1996learning,wong2007learning,zettlemoyer2007online,lu2008generative}. One prominent approach to build a semantic parser from scratch follows this procedure~\citep{wang2015building}:
\begin{enumerate}
\setlength{\parsep}{0pt}
\setlength{\parskip}{0pt}
    \item[a).] (canonical utterance, logical form) pairs are automatically generated according to a domain-general grammar and a domain-specific lexicon.
    \item[b).] Researchers use crowdsourcing to paraphrase those canonical utterances into natural language utterances~(the upper part of Figure \ref{fig:two-stage}).
    \item[c).] A semantic parser is built upon collected (natural language utterance, logical form) pairs.
\end{enumerate}

Canonical utterances are pseudo-language utterances automatically generated from grammar rules, which can be understandable to people, but do not sound natural. Though effective, the paraphrasing paradigm suffers from two drawbacks: (1) dependency on nontrivial human labor and (2) low utilization of canonical utterances.

Annotators may struggle to understand the exact meanings of canonical utterances. Some canonical utterances even incur ambiguity, which enhances the difficulty of annotation. Furthermore, \citet{wang2015building} and \citet{herzig2019don} only exploit them during data collection. Once the semantic parsing dataset is constructed, canonical utterances are thrown away, which leads to insufficient utilization. While \citet{herzig2019don} and \citet{su2017cross} have reported the effectiveness of leveraging them as intermediate outputs, they experiment in a completely supervised way, where the human annotation is indispensable.

In this work, inspired by unsupervised neural machine translation~\citep{lample2017unsupervised,artetxe2017unsupervised}, we propose a two-stage semantic parsing framework. The first stage uses a paraphrase model to convert natural language utterances into corresponding canonical utterances. The paraphrase model is trained in an \emph{unsupervised} way. Then a naive\footnote{We use word ``naive" just to differentiate from traditional semantic parser, where our module expects to accept canonical utterances instead of natural language utterances.} neural semantic parser is built upon auto-generated (canonical utterance, logical form) pairs using traditional \emph{supervised} training. These two models are concatenated into a pipeline~(Figure \ref{fig:two-stage}).

Paraphrasing aims to perform semantic normalization and reduce the diversity of expression, trying to bridge the gap between natural language and logical forms. The naive neural semantic parser learns inner mappings between canonical utterances and logical forms, as well as the structural constraints.

The unsupervised paraphrase model consists of one shared encoder and two separate decoders for natural language and canonical utterances. In the pre-training phase, we design three types of noise~(Section \ref{sec:noisy}) tailored for sentence-level denoising autoencoder~\citep{vincent2008extracting} task to warm up the paraphrase model without any parallel data. This task aims to reconstruct the raw input utterance from its corrupted version. After obtaining a good initialization point, we further incorporate back-translation~\citep{sennrich2015neural} and dual reinforcement learning~(Section \ref{sec:cycle_learning}) tasks during the cycle learning phase. In this phase, one encoder-decoder model acts as the environment to provide pseudo-samples and reward signals for another.

We conduct extensive experiments on benchmarks \textsc{Overnight} and \textsc{GeoGranno}, both in unsupervised and semi-supervised settings. The results show that our method obtains significant improvements over various baselines in unsupervised settings. With full labeled data, we achieve new state-of-the-art performances~($80.1\%$ on \textsc{Overnight} and $74.5\%$ on \textsc{GeoGranno}), not considering additional data sources.

The main contributions of this work can be summarized as follows:
\begin{itemize}
    \item A two-stage semantic parser framework is proposed, which casts parsing into paraphrasing. No supervision is provided in the first stage between input natural language utterances and intermediate output canonical utterances.
    \item In unsupervised settings, experimental results on datasets \textsc{Overnight} and \textsc{GeoGranno} demonstrate the superiority of our model over various baselines, including the supervised method in \citet{wang2015building} on \textsc{Overnight}~($60.7\%$ compared to $58.8\%$).
    \item The framework is also compatible with traditional supervised training and achieves the new state-of-the-art performances on datasets \textsc{Overnight} ($80.1\%$) and \textsc{GeoGranno} ($74.5\%$) with full labeled  data.
\end{itemize}

%% file: training.tex
\section{Our Approach}
\label{sec:training}

\subsection{Problem Definition}
For the rest of our discussion, we use $x$ to denote natural language utterance, $z$ for canonical utterance, and $y$ for logical form. $\mathbf{\mathcal{X}}$, $\mathbf{\mathcal{Z}}$ and $\mathbf{\mathcal{Y}}$ represent the set of all possible natural language utterances, canonical utterances, and logical forms respectively. The underlying mapping function $f:\mathcal{Z}\longrightarrow \mathcal{Y}$ is dominated by grammar rules.

We can train a naive neural semantic parser $P_{nsp}$ using attention~\citep{luong2015effective} based Seq2Seq model~\citep{sutskever2014sequence}. The labeled samples $\{(z,y), z\in \mathcal{Z},y\in \mathcal{Y}\}$ can be automatically generated by recursively applying grammar rules. $P_{nsp}$ can be pre-trained and saved for later usage. 

As for the paraphrase model~(see Figure \ref{fig:two-stage}), it consists of one shared encoder $E$ and two independent decoders: $D_x$ for natural language utterances and $D_z$ for canonical utterances. The symbol $\circ$ denotes module composition. Detailed model implementations are omitted here since they are not the main focus~(Appendix A.1 for reference).

Given an input utterance $x\in\mathcal{X}$, the paraphrase model $D_z\circ E$ converts it into possible canonical utterance $\hat{z}=D_z\circ E(x)$; then $\hat{z}$ is passed into the pre-trained naive parser $P_{nsp}$ to obtain predicted logical form $\hat{y}=P_{nsp}\circ D_z\circ E(x)$. Another paraphrase model, $D_x\circ E$, is only used as an auxiliary tool during training.

\subsection{Unsupervised training procedures}
To train an unsupervised paraphrase model with no parallel data between $\mathbf{\mathcal{X}}$ and $\mathbf{\mathcal{Z}}$, we split the entire training procedure into two phases: pre-training and cycle learning. $D_x\circ E$ and $D_z\circ E$ are first pre-trained as denoising auto-encoders~(DAE). This initialization phase plays a significant part in accelerating convergence due to the ill-posed nature of paraphrasing tasks. Next, in the cycle learning phase, we employ both back-translation~(BT) and dual reinforcement learning~(DRL) strategies for self-training and exploration.

\subsubsection{Pre-training phase}
In this phase, we initialize the paraphrase model via the denoising auto-encoder task. All auxiliary models involved in calculating rewards~(see Section \ref{sec:reward}) are also pre-trained.
\paragraph{Denoising auto-encoder}
\begin{figure}[htbp]
    \centering
    \includegraphics[width=\columnwidth]{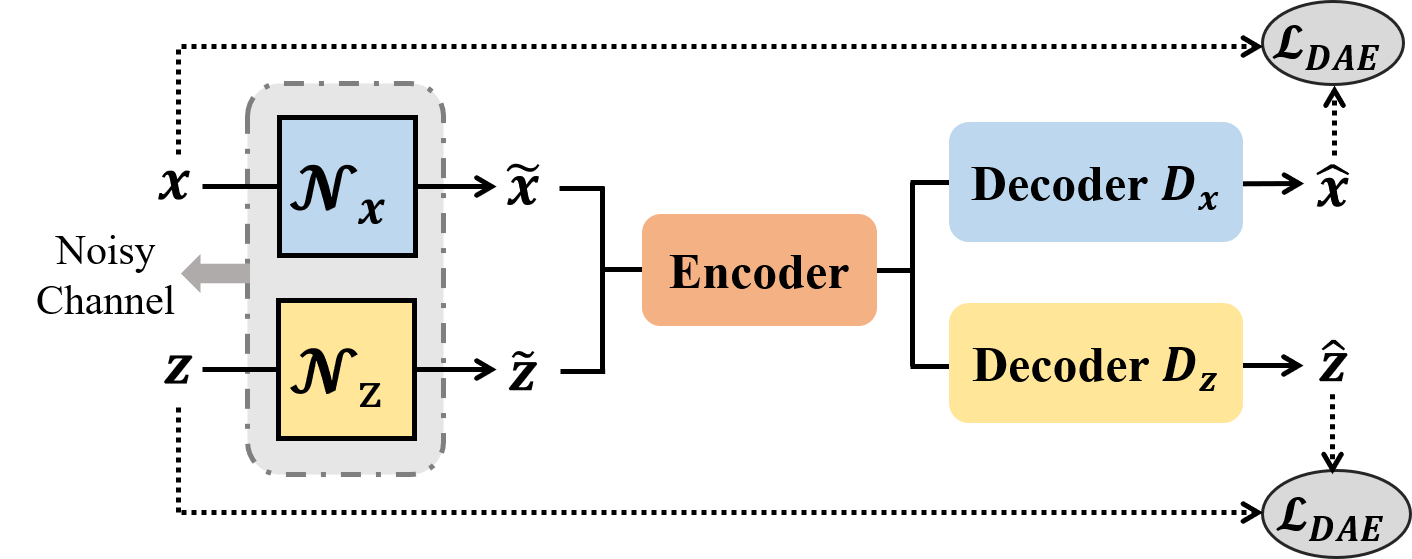}
    \caption{Denoising auto-encoders for natural language utterance $x$ and canonical utterance $z$.}
    \label{fig:dae}
\end{figure}
Given a natural language utterance $x$, we forward it through a noisy channel $\mathcal{N}_x(\cdot)$~(see Section \ref{sec:noisy}) and obtain its corrupted version $\tilde{x}$. Then, model $D_x\circ E$ tries to reconstruct the original input $x$ from its corrupted version $\tilde{x}$, see Figure \ref{fig:dae}. Symmetrically, model $D_z\circ E$ tries to reconstruct the original canonical utterance $z$ from its corrupted input $\mathcal{N}_z(z)$. The training objective can be formulated as
\begin{multline}\label{eq:dae}
\mathcal{L}_{DAE}=-\sum_{x\sim \mathcal{X}}\log P(x|\mathcal{N}_x(x);\Theta_{D_x\circ E})\\
-\sum_{z\sim\mathcal{Z}}\log P(z|\mathcal{N}_z(z);\Theta_{D_z\circ E})
\end{multline}
where $\Theta_{D_x\circ E}$ and $\Theta_{D_z\circ E}$ are parameters for the system.
\subsubsection{Cycle learning phase}
\label{sec:cycle_learning}
\begin{figure*}[htbp]
    \centering
    \includegraphics[width=.85\textwidth]{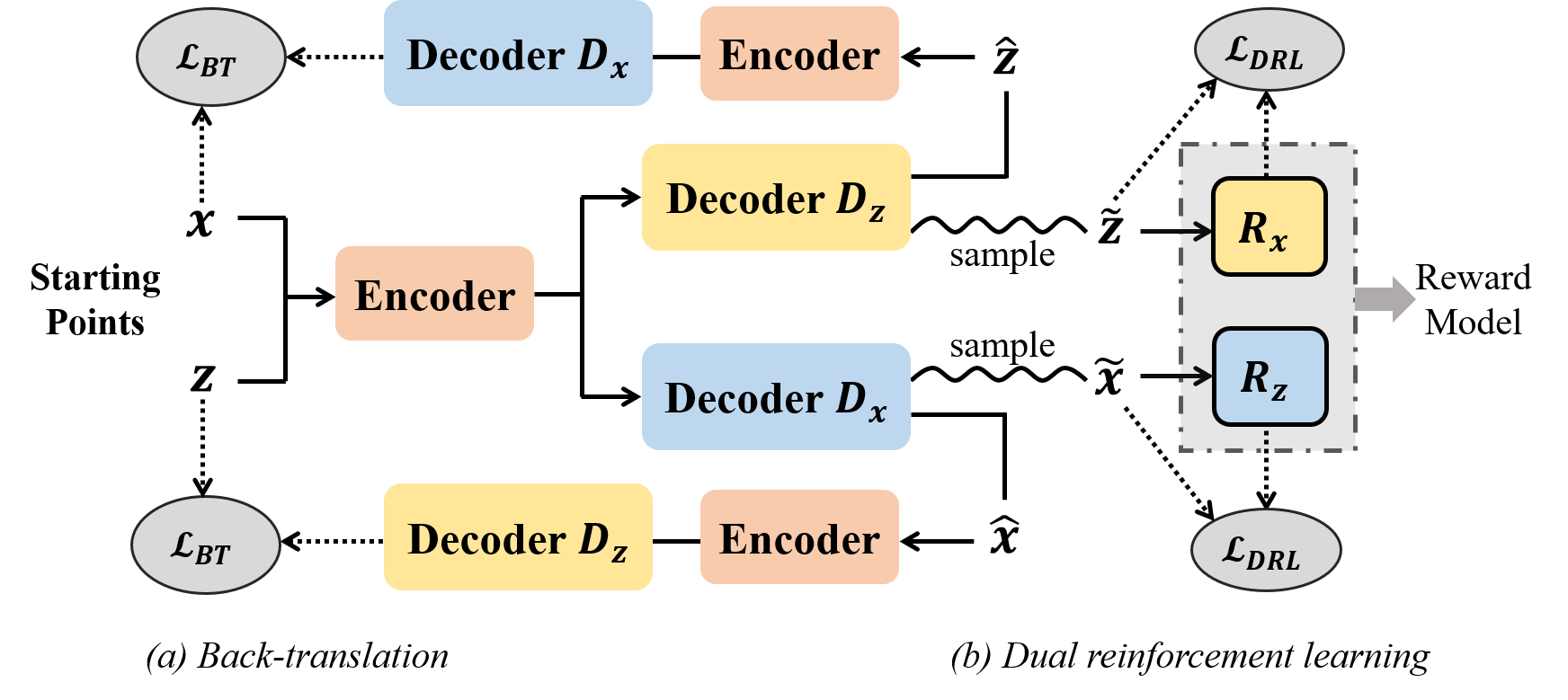}
    \caption{Cycle learning tasks: back-translation and dual reinforcement learning.}
    \label{fig:cycle}
\end{figure*}

The training framework till now is just a noisy-copying model. To improve upon it, we adopt two schemes in the cycle learning phase, back-translation~(BT) and dual reinforcement learning~(DRL), see Figure \ref{fig:cycle}.
\paragraph{Back-translation}
In this task, the shared encoder $E$ aims to map the input utterance of different types into the same latent space, and the decoders need to decompose this representation into the utterance of another type. More concretely, given a natural language utterance $x$, we use paraphrase model $D_{z}\circ E$ in evaluation mode with greedy decoding to convert $x$ into canonical utterance $\hat{z}$. We will obtain pseudo training sample $(\hat{z}, x)$ for paraphrase model $D_x\circ E$. Similarly, $(\hat{x},z)$ pair can be synthesized from model $D_x\circ E$ given canonical utterance $z$. Next, we train the paraphrase model from these pseudo-parallel samples and update parameters by minimizing
\begin{multline}\label{eq:bt}
    \mathcal{L}_{BT}=-\sum_{x\sim\mathcal{X}}\log P(x|D_z\circ E(x);\Theta_{D_x\circ E})\\
    - \sum_{z\sim\mathcal{Z}}\log P(z|D_x\circ E(z);\Theta_{D_z\circ E})
\end{multline}
 The updated model will generate better paraphrases during the iterative process.
\paragraph{Dual reinforcement learning}
Back-translation pays more attention to utilize what has been learned by the dual model, which may lead to a local optimum. To encourage more trials during cycle learning, we introduce the dual reinforcement learning strategy and optimize the system through policy gradient~\citep{sutton2000policy}.

Starting from a natural language utterance $x$, we sample one canonical utterance $\tilde{z}$ through $D_z\circ E$. Then, we evaluate the quality of $\tilde{z}$ from different aspects~(see Section \ref{sec:reward}) and obtain reward $R_x(\tilde{z})$. Similarly, we calculate reward $R_z(\tilde{x})$ for sampled natural language utterance $\tilde{x}$. To cope with high variance in reward signals, we increase sample size to $K$ and re-define reward signals via a baseline $b(\cdot)$ to stabilize learning: (take $\tilde{z}^k$ for an example)
\begin{align*}
R_x(\tilde{z}^k)\doteq R_x(\tilde{z}^k)-b(\tilde{z}),\quad k=1,\cdots,K
\end{align*}
We investigate different baseline choices~(such as running mean, cumulative mean of history, and reward of the greedy decoding prediction), and it performs best when we use the average of rewards within samples of per input, especially with larger sample size. The training objective is the negative sum of expected reward:
\begin{multline}
	\mathcal{L}_{DRL}=-\sum_{x\sim\mathcal{X}}\sum_{\tilde{z}} P(\tilde{z}|x;\Theta_{D_z\circ E})\cdot R_x(\tilde{z})\\
	-\sum_{z\sim\mathcal{Z}}\sum_{\tilde{x}} P(\tilde{x}|z;\Theta_{D_x\circ E})\cdot R_z(\tilde{x})\label{eq:drl}
\end{multline}
The gradient is calculated with REINFORCE~\citep{williams1992simple} algorithm:
\begin{multline*}
	\nabla \mathcal{L}\approx -\sum_{x\sim\mathcal{X}}\sum_{k}\frac{R_x(\tilde{z}^k)}{K} \nabla \log P(\tilde{z}^k|x;\Theta_{\tiny D_z\circ E})\\
	-\sum_{z\sim\mathcal{Z}}\sum_{k}\frac{R_z(\tilde{x}^k)}{K} \nabla\log P(\tilde{x}^k|z;\Theta_{D_x\circ E})
\end{multline*}

The complete loss function in the cycle learning phase is the sum of cross entropy loss and policy gradient loss: $\mathcal{L}_{Cycle}=\mathcal{L}_{BT}+\mathcal{L}_{DRL}$. The entire training procedure is summarized in Algorithm \ref{alg:train}.
\input{alg/alg.tex}

%% file: alg/alg.tex
\begin{algorithm}[htp]
\caption{Training procedure}
\label{alg:train}
\begin{algorithmic}[1]
\Require Unlabeled dataset $\mathcal{X},\mathcal{Z}$; Labeled $(z,y)$ pairs synthesized from grammar; Iterations $M$
\Ensure Paraphrase model $D_z^{(M)}\circ E^{(M)}$

{\color{blue}\Comment{Pre-training phase}}
\State{Pre-train all auxiliary models: language models $LM_x$ and $LM_z$, naive neural semantic parser $P_{nsp}$ and utterance discriminator $P_{dis}$}
\State{Pre-train paraphrase models $D_x^{(0)}\circ E^{(0)}$ and $D_z^{(0)}\circ E^{(0)}$ via objective $\mathcal{L}_{DAE}$ based on Eq.\ref{eq:dae}}

{\color{blue}\Comment{Cycle learning phase}}
\For{$i=0$ to $M-1$}
\State{Sample natural language utterance $x\sim\mathcal{X}$}
\State{Sample canonical utterance $z\sim\mathcal{Z}$}

{\color{blue}\Comment{Back-translation}}
\State{Generate $\hat{z}$ via model $D_z^{(i)}\circ E^{(i)}(x)$;
\State{Generate $\hat{x}$ via model $D_x^{(i)}\circ E^{(i)}(z)$;}
}
\State{Use $(\hat{z},x)$ and $(\hat{x},z)$ as pseudo samples, calculate loss $\mathcal{L}_{BT}$ based on Eq.\ref{eq:bt};}

{\color{blue}\Comment{Dual Reinforcement Learning}}
\State{Sample $\tilde{z}$ via model $D_z^{(i)}\circ E^{(i)}(x)$}
\State{Compute total reward $R_x(\tilde{z})$ via models $LM_z$, $P_{dis}$, $P_{nsp}$ and $D_x^{(i)}\circ E^{(i)}$ based on Eq.\ref{eq:x}}
\State{Sample $\tilde{x}$ via model $D_x^{(i)}\circ E^{(i)}(z)$}
\State{Compute total reward $R_z(\tilde{x})$ via models $LM_x$, $P_{dis}$ and $D_z^{(i)}\circ E^{(i)}$ based on Eq.\ref{eq:z}}
\State{Given $R_x(\tilde{z})$ and $R_z(\tilde{x})$, calculate loss $\mathcal{L}_{DRL}$ based on Eq.\ref{eq:drl}}

{\color{blue}\Comment{Update model parameters}}
\State{Calculate total loss $\mathcal{L}_{Cycle}=\mathcal{L}_{BT}+\mathcal{L}_{DRL}$}
\State{Update model parameters, get new models $D_x^{(i+1)}\circ E^{(i+1)}$ and $D_z^{(i+1)}\circ E^{(i+1)}$}
\EndFor
\State{\Return $D_z^{(M)}\circ E^{(M)}$}
\end{algorithmic}
\end{algorithm}

%% file: details.tex
\section{Training details}
\label{sec:detail}
In this section, we elaborate on different types of noise used in our experiment and the reward design in dual reinforcement learning.

\subsection{Noisy channel}
\label{sec:noisy}
We introduce three types of noise to deliberately corrupt the input utterance in the DAE task.

\paragraph{Importance-aware word dropping}
Traditional word dropping~\citep{lample2017unsupervised} discards each word in the input utterance with equal probability $p_{wd}$. During reconstruction, the decoder needs to recover those words based on the context. We further inject a bias towards dropping more frequent words~(such as function words) in the corpus instead of less frequent words~(such as content words), see Table \ref{tab:drop} for illustration. 
\begin{table}[htbp]
    \centering{\small
    \begin{tabular}{|c|l|}\hline
        Input $x$ & {\small \textbf{what} team \textbf{does} kobe bryant play for} \\\hline
        Ordinary drop & {\small what does kobe bryant for} \\\hline
        Our drop & {\small team kobe bryant play for}\\\hline
    \end{tabular}}
    \caption{Importance-aware word dropping example.}
    \label{tab:drop}
\end{table}

Each word $x_i$ in the natural language utterance $x=(x_1,x_2,\cdots,x_{|x|})$ is independently dropped with probability
$$p_{wd}(x_i)=min\{p_{max},w(x_i)/\sum_{j=1}^{|x|}w(x_j)\}$$
where $w(x_i)$ is the word count of $x_i$ in $\mathcal{X}$, and $p_{max}$ is the maximum dropout rate~($p_{max}=0.2$ in our experiment). As for canonical utterances, we apply this word dropping similarly.

\paragraph{Mixed-source addition}
For any given raw input, it is either a natural language utterance or a canonical utterance. This observation discourages the shared encoder $E$ to learn a common representation space. Thus, we propose to insert extra words from another source into the input utterance. As for noisy channel $\mathcal{N}_x(\cdot)$, which corrupts a natural language utterance, we first select one candidate canonical utterance $z$; next, $10\%$-$20\%$ words are randomly sampled from $z$ and inserted into arbitrary position in $x$, see Table \ref{tab:add} for example.

To pick candidate $z$ with higher relevance, we use a heuristic method: $C$ canonical utterances are randomly sampled as candidates~($C=50$); we choose $z$ that has the minimum amount of Word Mover's Distance concerning $x$~(WMD,~\citealp{kusner2015word}). The additive operation is exactly symmetric for noisy channel $\mathcal{N}_z$.
\begin{table}[htbp]
    \centering
    \begin{tabular}{|c|l|}\hline
        Input $x$ & {\small how many players are there} \\\hline
        Selected $z$ & {\small \textbf{number} of team} \\\hline
        Output $\tilde{x}$ & {\small how many \textbf{number} players are there}\\\hline
    \end{tabular}
    \caption{Mixed-source addition example.}
    \label{tab:add}
\end{table}
\paragraph{Bigram shuffling}
We also use word shuffling~\citep{lample2017unsupervised} in noisy channels. It has been proven useful in preventing the encoder from relying too much on the word order. Instead of shuffling words, we split the input utterance into n-grams first and shuffle at n-gram level~(bigram in our experiment). Considering the inserted words from another source, we shuffle the entire utterance after the addition operation~(see Table \ref{tab:shuffle} for example).
\begin{table}[htbp]
    \centering
    \begin{tabular}{|c|l|}\hline
        Input $x$ & {\small what is kobe bryants team} \\\hline
        1-gram shuffling & {\small what is kobe team \textbf{bryants}} \\\hline
        2-gram shuffling & {\small what is team \textbf{kobe bryants}} \\\hline
    \end{tabular}
    \caption{Bigram shuffling example}
    \label{tab:shuffle}
\end{table}

\subsection{Reward design}
\label{sec:reward}
In order to provide more informative reward signals and promote the performance in the DRL task, we introduce various rewards from different aspects.
\paragraph{Fluency}
The fluency of an utterance is evaluated by a length-normalized language model. We use individual language models~($LM_x$ and $LM_z$) for each type of utterances. As for a sampled natural language utterance $\tilde{x}$, the fluency reward is
\begin{align*}
R^{flu}_z(\tilde{x})=\frac{1}{|\tilde{x}|}\log LM_x(\tilde{x})
\end{align*}
As for canonical utterances, we also include an additional $0/1$ reward from downstream naive semantic parser to indicate whether the sampled canonical utterance $\tilde{z}$ is well-formed as input for $P_{nsp}$. 
\begin{align*}
\hat{y}=&\underset{y}{\arg max}P_{nsp}(y|\tilde{z}), \text{ greedy decoding}\\
\begin{split}
R_x^{flu}(\tilde{z})=&\frac{1}{|\tilde{z}|}\log LM_z(\tilde{z})\\
&+\mathbb{I}\cdot\{\text{no error while executing }\hat{y}\}
\end{split}
\end{align*}

\paragraph{Style}
Natural language utterances are diverse, casual, and flexible, whereas canonical utterances are generally rigid, regular, and restricted to some specific form induced by grammar rules. To distinguish their characteristics, we incorporate another reward signal that determine the style of the sampled utterance. This is implemented by a CNN discriminator~\citep{kim2014convolutional}:
\begin{align*}
R^{sty}_z(\tilde{x})=1-P_{dis}(\tilde{x});\quad R^{sty}_x(\tilde{z})=P_{dis}(\tilde{z})
\end{align*}
where $P_{dis}(\cdot)$ is a pre-trained sentence classifier that evaluates the probability of the input utterance being a canonical utterance.
\paragraph{Relevance}
Relevance reward is included to measure how much content is preserved after paraphrasing. We follow the common practice to take the loglikelihood from the dual model.
\begin{align*}
R_x^{rel}(x, \tilde{z})=&\log P(x|\tilde{z};\Theta_{D_{x}\circ E}) \\
R_z^{rel}(z, \tilde{x})=&\log P(z|\tilde{x};\Theta_{D_{z}\circ E})
\end{align*}
Some other methods include computing the cosine similarity of sentence vectors or BLEU score~\citep{papineni2002bleu} between the raw input and the reconstructed utterance. Nevertheless, we find loglikelihood to perform better in our experiments.

The total reward for the sampled canonical utterance $\tilde{z}$ and natural language utterance $\tilde{x}$ can be formulated as
\begin{align}
R_x(\tilde{z})=&R_x^{flu}(\tilde{z})+R_x^{sty}(\tilde{z})+R_x^{rel}(x,\tilde{z})\label{eq:x} \\
R_z(\tilde{x})=&R_z^{flu}(\tilde{x})+R_z^{sty}(\tilde{x})+R_z^{rel}(z,\tilde{x})\label{eq:z}
\end{align}

%% file: experiment.tex
\section{Experiment}
\label{sec:experiment}
In this section, we evaluate our system on benchmarks \textsc{Overnight} and \textsc{GeoGranno} in both unsupervised and semi-supervised settings. Our implementations are public available\footnote{\url{https://github.com/rhythmcao/unsup-two-stage-semantic-parsing}}.
\paragraph{\textsc{Overnight}}
 It contains natural language paraphrases paired with logical forms over $8$ domains. We follow the traditional $80\%/20\%$ train/valid to choose the best model during training. Canonical utterances are generated with tool \textsc{Sempre}\footnote{\url{https://github.com/percyliang/sempre}} paired with target logical forms~\citep{wang2015building}. Due to the limited number of grammar rules and its coarse-grained nature, there is only one canonical utterance for each logical form, whereas $8$ natural language paraphrases for each canonical utterance on average. For example, to describe the concept of ``larger", in natural language utterances, many synonyms, such as ``more than", ``higher", ``at least", are used, while in canonical utterances, the expression is restricted by grammar.
\paragraph{\textsc{GeoGranno}} Due to the language mismatch problem~\citep{herzig2019don}, annotators are prone to reuse the same phrase or word while paraphrasing. \textsc{GeoGranno} is created via detection instead of paraphrasing. Natural language utterances are firstly collected from query logs. Crowd workers are required to select the correct canonical utterance from candidate list~(provided by an incrementally trained score function) per input. We follow exactly the same split~(train/valid/test $487/59/278$) in original paper \citet{herzig2019don}.

\subsection{Experiment setup}
Throughout the experiments, unless otherwise specified, word vectors are initialized with Glove6B~\cite{pennington2014glove} with $93.3\%$ coverage on average and allowed to fine-tune. Out-of-vocabulary words are replaced with $\left<unk\right>$.  Batch size is fixed to $16$ and sample size $K$ in the DRL task is $6$. During evaluation, the size of beam search is $5$. We use optimizer Adam~\citep{kingma2014adam} with learning rate $0.001$ for all experiments. All auxiliary models are pre-trained and fixed for later usage. We report the denotation-level accuracy of logical forms in different settings.

\paragraph{Supervised settings}
This is the traditional scenario, where labeled $(x,y)$ pairs are used to train a one-stage parser directly, $(x,z)$ and $(z,y)$ pairs are respectively used to train different parts of a two-stage parser.

\paragraph{Unsupervised settings}
We split all methods into two categories: one-stage and two-stage. In the one-stage parser, \textsc{Embed} semantic parser is merely trained on $(z,y)$ pairs but evaluated on natural language utterances. Contextual embeddings ELMo~\citep{peters2018deep} and Bert-base-uncased~\citep{devlin2018bert} are also used to replace the original embedding layer; \textsc{WmdSamples} method labels each input $x$ with the most similar logical form~(one-stage) or canonical utterance~(two-stage) based on WMD~(\citealp{kusner2015word}) and deals with these faked samples in a supervised way; \textsc{MultiTaskDae} utilizes another decoder for natural language utterances in one-stage parser to perform the same DAE task discussed before. The two-stage \textsc{CompleteModel} can share the encoder or not~(\textsc{-SharedEncoder}), and include tasks in the cycle learning phase or not~(\textsc{-CycleLearning}). The downstream parser $P_{nsp}$ for the two-stage system is \textsc{Embed + Glove6B} and fixed after pre-training.
\paragraph{Semi-supervised settings}
To further validate our framework, based on the complete model in unsupervised settings, we also conduct semi-supervised experiments by gradually adding part of labeled paraphrases with supervised training into the training process~(both pre-training and cycle learning phase).

\subsection{Results and analysis}
\input{tab/main.tex}
\input{tab/main_geo}
As Table \ref{tab:main} and \ref{tab:main_geo} demonstrate, in unsupervised settings: (1) two-stage semantic parser is superior to one-stage, which bridges the vast discrepancy between natural language utterances and logical forms by utilizing canonical utterances. Even in supervised experiments, this pipeline is still competitive~($76.4\%$ compared to $76.0\%$, $71.6\%$ to $71.9\%$). (2) Not surprisingly, model performance is sensitive to the word embedding initialization. On \textsc{Overnight}, directly using raw Glove6B word vectors, the performance is the worst among all baselines~($19.7\%$). Benefiting from pre-trained embeddings ELMo or Bert, the accuracy is dramatically improved~($26.2\%$ and $32.7\%$). (3) When we share the encoder module in a one-stage parser for multi-tasking~(\textsc{MultiTaskDae}), the performance is not remarkably improved, even slightly lower than \textsc{Embed+Bert}~($31.9\%$ compared to $32.7\%$, $38.1\%$ to $40.7\%$). We hypothesize that a semantic parser utilizes the input utterance in a way different from that of a denoising auto-encoder, thus focusing on different zones in representation space. However, in a paraphrasing model, since the input and output utterances are exactly symmetric, sharing the encoder is more suitable to attain an excellent performance~(from $57.5\%$ to $60.7\%$ on \textsc{Overnight}, $59.0\%$ to $63.7\%$ on \textsc{GeoGranno}). Furthermore, the effectiveness of the DAE pre-training task~($44.9\%$ and $44.6\%$ accuracy on target task) can be explained in part by the proximity of natural language and canonical utterances. (4) \textsc{WmdSamples} method is easy to implement but has poor generalization and obvious upper bound. While our system can self-train through cycle learning and promote performance from initial $44.9\%$ to $60.7\%$ on \textsc{Overnight}, outperforming traditional supervised method~\citep{wang2015building} by $1.9$ points.

As for semi-supervised results: (1) when only $5\%$ labeled data is added, the performance is dramatically improved from $60.7\%$ to $68.4\%$ on \textsc{Overnight} and $63.7\%$ to $69.4\%$ on \textsc{GeoGranno}. (2) With $30\%$ annotation, our system is competitive~($75.0\%/71.6\%$) to the neural network model using all data with supervised training. (3) Compared with the previous result reported in \citet{cao-etal-2019-semantic} on dataset \textsc{Overnight} with $50\%$ parallel data, our system surpasses it by a large margin~($4\%$) and achieves the new state-of-the-art performance on both datasets when using all labeled data~($80.1\%/74.5\%$), not considering results using additional data sources or cross-domain benefits. 

\begin{figure}[htbp]
    \centering
    \includegraphics[width=0.8\columnwidth]{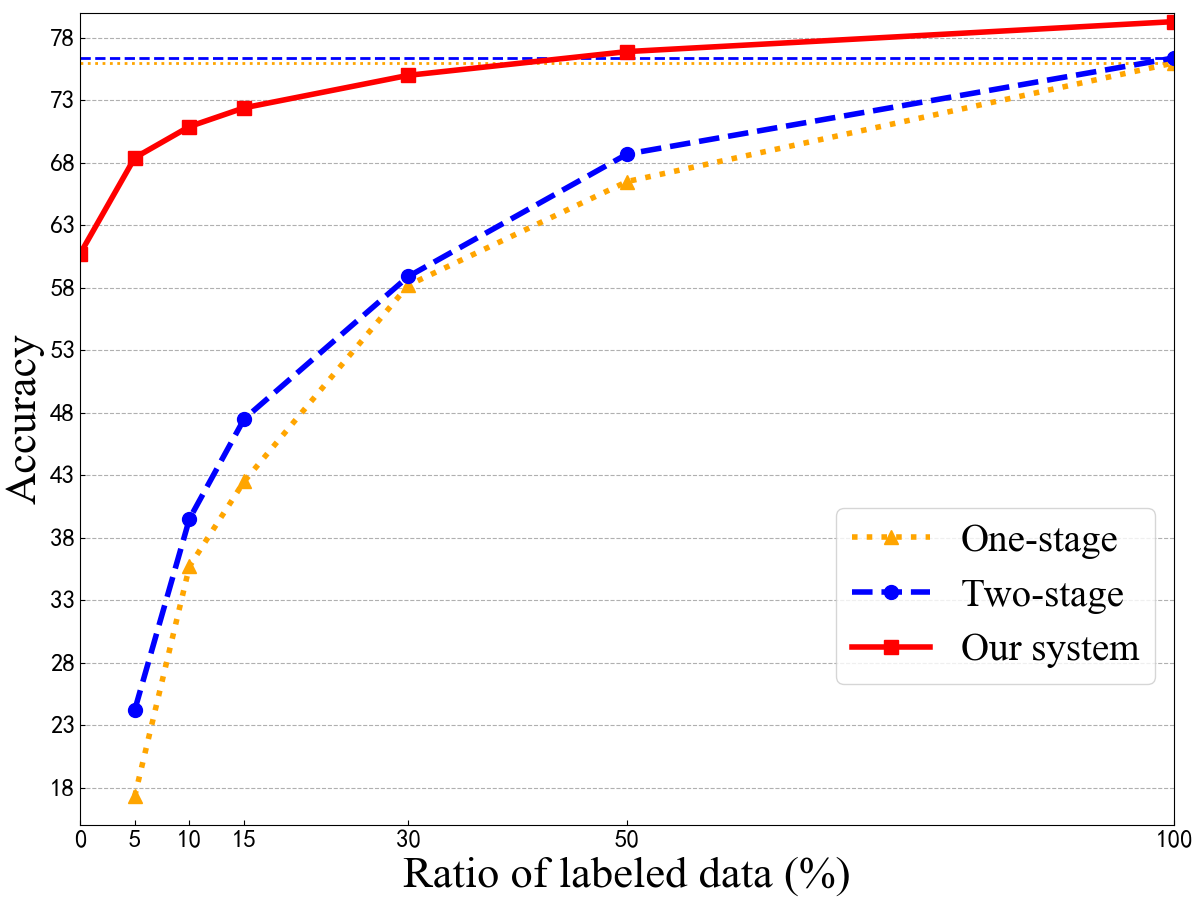}
    \caption{Semi-supervised results of different ratios of labeled data on \textsc{Overnight}. Baselines are one-stage and two-stage models with merely supervised training.}
    \label{fig:semi}
\end{figure}

From the experimental results and Figure \ref{fig:semi}, we can safely summarize that (1) our proposed method resolves the daunting problem of cold start when we train a semantic parser without any parallel data. (2) It is also compatible with traditional supervised training and can easily scale up to handle more labeled data.

\subsection{Ablation study}
In this section, we analyze the influence of each noise type in the DAE task and different combinations of schemes in the cycle learning phase on dataset \textsc{Overnight}.
\subsubsection{Noisy channels in the pre-training DAE}
\input{tab/noise.tex}
According to results in Table \ref{tab:noise}, (1) it is interesting that even without any noise, in which case the denoising auto-encoder degenerates into a simple copying model, the paraphrase model still succeeds to make some useful predictions~($26.9\%$). This observation may be attributed to the shared encoder for different utterances. (2) When we gradually complicate the DAE task by increasing the number of noise types, the generalization capability continues to improve. (3) Generally speaking, importance-aware drop and mixed-source addition are more useful than bigram shuffling in this task.
\subsubsection{Strategies in the cycle learning}
\input{tab/cycle.tex}
The most striking observation arising from Table \ref{tab:cycle} is that the performance decreases by $1.5$ percent when we add the DAE task into the cycle learning phase~(BT+DRL). A possible explanation for this phenomenon may be that the model has reached its bottleneck of the DAE task after pre-training, thereby making no contribution to the cycle learning process. Another likely factor may stem from the contradictory goals of different tasks. If we continue to add this DAE regularization term, it may hinder exploratory trials of the DRL task. By decoupling the three types of rewards in DRL, we discover that style and relevance rewards are more informative than the fluency reward.

\subsection{Case study}
\input{tab/case.tex}
In Table \ref{tab:case}, we compare intermediate canonical utterances generated by our unsupervised paraphrase model with that created by the baseline \textsc{WmdSamples}. In domain \textsc{Basketball}, our system succeeds in paraphrasing the constraint into ``{\it at least 3}'', which is an alias of ``{\it 3 or more}''. This finding consolidates the assumption that our model can learn these fine-grained semantics, such as phrase alignments. In domain \textsc{GeoGranno}, our model rectifies the errors in baseline system where constraint ``{\it borders \_state\_}" is missing and subject ``{\it state}" is stealthily replaced with ``{\it population}". As for domain \textsc{Calendar}, the baseline system fails to identify the query object and requires ``{\it meeting}'' instead of ``{\it person}''. Although our model correctly understands the purpose, it is somewhat stupid to do unnecessary work. The requirement ``{\it attendee of weekly standup}'' is repeated. This may be caused by the uncontrolled process during cycle learning in that we encourage the model to take risky steps for better solutions.

%% file: tab/main.tex
\begin{table*}[ht]
  \setlength\aboverulesep{0pt}
  \setlength\belowrulesep{0pt}
  \centering{\small
  
  \begin{adjustbox}{width=\textwidth}
    \begin{tabular}{c|p{14.6em}|cccccccc|c}
    \toprule
    \multicolumn{2}{c|}{\textbf{Method}} & \textbf{Bas} & \textbf{Blo} & \textbf{Cal} & \textbf{Hou} & \textbf{Pub} & \textbf{Rec} & \textbf{Res} & \textbf{Soc} & \textbf{Avg} \\
    \toprule
    \multicolumn{11}{c}{\textbf{Supervised}} \\
    \hline
    \multirow{6}[2]{*}{Previous} & SPO~\citep{wang2015building} & 46.3  & 41.9  & 74.4  & 54.0  & 59.0  & 70.8  & 75.9  & 48.2  & 58.8  \\
      & DSP-C~\citep{xiao-etal-2016-sequence} & 80.5  & 55.6  & 75.0  & 61.9  & 75.8  & $\_$ & 80.1  & 80.0  & 72.7  \\
      & \textsc{NoRecomb}*~\citep{jia-liang-2016-data} & 85.2  & 58.1  & 78.0  & 71.4  & 76.4  & 79.6  & 76.2  & 81.4  & 75.8  \\
     & \textsc{CrossDomain}*~\citep{su2017cross} & 86.2 & 60.2 & 79.8 & 71.4 & 78.9 & 84.7 & 81.6 & 82.9 & 78.2 \\ 
     & \textsc{Seq2Action}~\citep{chen2018sequence} & \textbf{88.2} & 61.4 & \textbf{81.5} & 74.1 & 80.7 & 82.9 & 80.7 & 82.1 & 79.0 \\
     & \textsc{Dual}*~\citep{cao-etal-2019-semantic} & 87.5 & 63.7 & 79.8 & 73.0 & \textbf{81.4} & 81.5 & 81.6 & \textbf{83.0} & 78.9 \\
    \hline
    \multirow{2}[2]{*}{Ours} & One-stage & 85.2  & 61.9  & 73.2  & 72.0  & 76.4  & 80.1  & 78.6  & 80.8  & 76.0  \\
      & Two-stage & 84.9  & 61.2  & 78.6  & 67.2  & 78.3  & 80.6  & 78.9  & 81.3  & 76.4 \\
    \toprule
    \multicolumn{11}{c}{\textbf{Unsupervised}} \\
    \hline
    \multirow{5}{*}{One-stage} & \textsc{Embed + Glove6B} & 22.3  & 23.6  & 9.5  & 26.5  & 18.0  & 24.5  & 24.7  & 8.4  & 19.7  \\
      & \textsc{\qquad\quad\ \ + ELMo}  & 36.8  & 21.1  & 20.2  & 21.2  & 23.6  & 36.1  & 37.7  & 12.8  & 26.2  \\
      & \textsc{\qquad\quad\ \ + Bert}  & 40.4  & 31.6 & 23.2  & 35.5  & 37.9 & 30.1 & 44.0  & 19.2  &  32.7 \\
      & \textsc{WmdSamples} & 34.5  & 33.8  & 29.2  & 37.6  & 36.7  & 41.7  & 56.6  & 37.0  & 38.4  \\
      & \textsc{MultiTaskDae} & 44.0  & 25.8  & 16.1  & 34.4  & 29.2  & 46.3  & 43.7  & 15.5  & 31.9  \\
    \hline
    \multirow{5}{*}{Two-stage} & \textsc{WmdSamples} & 31.9  & 29.0  & 36.1  & 47.9  & 34.2  & 41.0  & 53.8  & 35.8  & 38.7  \\
      & \textsc{CompleteModel}  & \textbf{64.7}  & \textbf{53.4}  & 58.3  & 59.3  & \textbf{60.3}  & \textbf{68.1}  & \textbf{73.2}  & \textbf{48.4}  & \textbf{60.7}  \\
      & \textsc{\quad - CycleLearning} & 32.5  & 43.1  & 36.9  & 48.2  & 53.4  & 49.1  & 58.7  & 36.9  & 44.9  \\
      & \textsc{\quad - SharedEncoder}  & 63.4  & 46.4  & \textbf{58.9}  & \textbf{61.9}  & 56.5  & 65.3  & 64.8  & 42.9  & 57.5  \\
    \toprule
     \multicolumn{11}{c}{\textbf{Semi-supervised}} \\\hline
     \multicolumn{2}{l|}{\textsc{Dual}~\citep{cao-etal-2019-semantic} + $50\%$ labeled data} & 83.6  & 62.2  & 72.6  & 61.9  & 71.4  & 75.0  & 76.5  & 80.4  & 73.0 \\\hline
     \multicolumn{2}{l|}{\textsc{CompleteModel} + $5\%$ labeled data} & 83.6 & 57.4 & 66.1 & 63.0 & 60.3 & 68.1 & 75.3 & 73.1 & 68.4 \\
     \multicolumn{2}{l|}{\text{\qquad\qquad\qquad\qquad} + $15\%$ labeled data} & 84.4 & 59.4 & 79.2 & 57.1 & 65.2 & 79.2 & 77.4 & 76.9 & 72.4 \\
     \multicolumn{2}{l|}{\text{\qquad\qquad\qquad\qquad} + $30\%$ labeled data} & 85.4 & 64.9 & 77.4 & 69.3 & 67.1 & 78.2 & 79.2 & 78.3 & 75.0 \\
     \multicolumn{2}{l|}{\text{\qquad\qquad\qquad\qquad} + $50\%$ labeled data} & 85.9 & 64.4 & \textbf{81.5} & 66.1 & 74.5 & 82.4 & 79.8 & 81.6 & 77.0 \\
     \multicolumn{2}{l|}{\text{\qquad\qquad\qquad\qquad} + $100\%$ labeled data} & 87.2 & \textbf{65.7} & 80.4 & \textbf{75.7} & 80.1 & \textbf{86.1} & \textbf{82.8} & 82.7 & \textbf{80.1} \\\bottomrule
    \end{tabular}%
    \end{adjustbox}
}
  \caption{Denotation level accuracy of logical forms on dataset \textsc{Overnight}. Previous supervised methods with superscript * means cross-domain or extra data sources are not taken into account.}
  \label{tab:main}%
\end{table*}%

%% file: tab/main_geo.tex
\begin{table}[ht]
  \setlength\aboverulesep{0pt}
  \setlength\belowrulesep{0pt}
  \centering{\small
  \begin{adjustbox}{width=0.48\textwidth}
    \begin{tabular}{c|p{10em}|c}
    \toprule
    \multicolumn{2}{c|}{\textbf{Method}} & \textbf{\textsc{GeoGranno}} \\
    \toprule
    \multicolumn{3}{c}{\textbf{Supervised}} \\
    \hline
    \multirow{2}{*}{Previous} & \textsc{CopyNet+ELMo}~\citep{herzig2019don} & \multirow{2}{*}{72.0} \\
    \hline
    \multirow{2}{*}{Ours} & One-stage & 71.9  \\
      & Two-stage & 71.6 \\
    \toprule
    \multicolumn{3}{c}{\textbf{Unsupervised}} \\
    \hline
    \multirow{5}{*}{One-stage} & \textsc{Embed + Glove6B} & 36.7 \\
      & \textsc{\qquad\quad\ \ + ELMo}  & 38.9 \\
      & \textsc{\qquad\quad\ \ + Bert}  & 40.7 \\
      & \textsc{WmdSamples} & 32.0 \\
      & \textsc{MultiTaskDae} & 38.1 \\
    \hline
    \multirow{5}{*}{Two-stage} & \textsc{WmdSamples} & 35.3 \\
      & \textsc{CompleteModel} & 63.7 \\
      & \textsc{\quad - CycleLearning} & 44.6 \\
      & \textsc{\quad - SharedEncoder} & 59.0 \\
    \toprule
     \multicolumn{3}{c}{\textbf{Semi-supervised}} \\\hline
     \multicolumn{2}{l|}{\textsc{CompleteModel} + $5\%$ labeled data} & 69.4 \\
     \multicolumn{2}{l|}{\text{\qquad\qquad\qquad\qquad} + $30\%$ labeled data} & 71.6\\
     \multicolumn{2}{l|}{\text{\qquad\qquad\qquad\qquad} + $100\%$ labeled data} & \textbf{74.5} \\\bottomrule
    \end{tabular}%
    \end{adjustbox}
}
  \caption{Denotation level accuracy of logical forms on dataset \textsc{GeoGranno}.}
  \label{tab:main_geo}%
\end{table}%

%% file: tab/noise.tex
\begin{table}[htbp]
  \centering \small
    \setlength\aboverulesep{0pt}
  \setlength\belowrulesep{0pt}
    \begin{tabular}{c|ccc|c}\toprule
   \textbf{\# Types} & \textbf{Drop} & \textbf{Addition} & \textbf{Shuffling} & \textbf{Acc} \\\hline
    none &  &  &  & 26.9  \\\hline
    \multirow{3}[1]{*}{one} & \checkmark &  &  & 33.7  \\
      &  & \checkmark &  & 32.1  \\
      &  &  & \checkmark & 31.6  \\
    \hline
    \multirow{3}[2]{*}{two} & \checkmark & \checkmark &  & 43.0  \\
      & \checkmark &  & \checkmark & 38.0  \\
      &  & \checkmark & \checkmark & 36.0  \\
    \hline
    all & \checkmark & \checkmark & \checkmark & \textbf{44.9}  \\
    \bottomrule
    \end{tabular}%

  \caption{Ablation study of different noisy channels.}
  \label{tab:noise}%
\end{table}%

%% file: tab/cycle.tex
\begin{table}[htbp]
  \centering\small
  \setlength\aboverulesep{0pt}
  \setlength\belowrulesep{0pt}
      \begin{tabular}{ccc|c}
    \toprule
    \textbf{DAE} & \textbf{BT} & \textbf{DRL} & \textbf{Acc}\\
    \hline
    \checkmark &  &  & 44.9  \\
     & \checkmark &  & 51.9  \\
     &  & \checkmark & 55.9  \\
    \checkmark & \checkmark &  & 53.2  \\
    \checkmark &  & \checkmark & 53.7  \\
     & \checkmark & \checkmark & \textbf{60.7 } \\
    \checkmark & \checkmark & \checkmark & 59.2 \\
    \bottomrule
    \end{tabular}%
  \caption{Ablation study of schemes in cycle learning}
  \label{tab:cycle}%
\end{table}%

%% file: tab/case.tex
\begin{table}[t]
    \centering \small
    \resizebox{\columnwidth}{!}{
    \begin{tabular}{p{\columnwidth}}
    \toprule
    \textbf{Input:} {\it who has gotten \textcolor{blue}{3 or more} steals}\\
    \textbf{Baseline:} player whose number of steals ( over a season ) is  \textcolor{red}{at most 3}\\
    \textbf{Ours:} player whose number of steals ( over a season ) is \textcolor{blue}{at least 3}\\
    \makecell[c]{(a) domain: \textsc{Basketball}} \\
    \midrule
    \textbf{Input:} {\it show me all \textcolor{blue}{attendees} of the weekly standup meeting}\\
    \textbf{Baseline:} \textcolor{red}{meeting} whose attendee is attendee of weekly standup\\
    \textbf{Ours:} \textcolor{blue}{person} that is attendee of weekly standup and that is attendee of weekly standup\\
    \makecell[c]{(b) domain: \textsc{Calendar}}\\
    \midrule
    \textbf{Input:} {\it what is the largest state \textcolor{blue}{bordering \_state\_}}\\
    \textbf{Baseline:} state that has the largest area\\
    \textbf{Ours:} state that \textcolor{blue}{borders \_state\_} and that has the largest area\\
    \\
    \textbf{Input:} {\it which \textcolor{blue}{state} has the highest population density ?}\\
    \textbf{Baseline:} \textcolor{red}{population} of state that has the largest density\\
    \textbf{Ours:} \textcolor{blue}{state} that has the largest density \\    
    \makecell[c]{(c) domain: \textsc{GeoGranno}}\\    
    \bottomrule
    \end{tabular}
    }
    \caption{Case study. The input is natural language utterance, and the intermediate output is canonical utterance. Entities in dataset \textsc{GeoGranno} are replaced with its types, e.g. ``\_state\_".}
    \label{tab:case}
\end{table}

%% file: related.tex
\section{Related Work}
\label{sec:related}

\paragraph{Annotation for Semantic Parsing}
Semantic parsing is always data-hungry. However, the annotation for semantic parsing is not user-friendly. Many researchers have attempted to relieve the burden of human annotation, such as training from weak supervision~\cite{krishnamurthy2012weakly,berant2013semantic,liang2016neural,goldman2017weakly}, semi-supervised learning~\citep{yin-etal-2018-structvae,guo2018question,cao-etal-2019-semantic,zhu2014semantic}, on-line learning~\citep{iyer2017learning,lawrence2018improving} and relying on multi-lingual~\citep{zou-lu-2018-learning} or cross-domain datasets~\citep{herzig2017neural,zhao-etal-2019-data}. In this work, we try to avoid the heavy work in annotation by utilizing canonical utterances as intermediate results and construct an unsupervised model for paraphrasing.

\paragraph{Unsupervised Learning for Seq2Seq Models}
Seq2Seq~\citep{sutskever2014sequence,zhu2017encoder} models have been successfully applied in unsupervised tasks such as neural machine translation~(NMT)~\citep{lample2017unsupervised, artetxe2017unsupervised}, text simplification~\citep{zhaosemi}, spoken language understanding~\citep{zhu2018robust} and text style transfer~\citep{DBLP:conf/ijcai/LuoLZYCSS19}. Unsupervised NMT relies heavily on pre-trained cross-lingual word embeddings for initialization, as \citet{lample-etal-2018-phrase} pointed out. Moreover, it mainly focuses on learning phrase alignments or word mappings. While in this work, we dive into sentence-level semantics and adopt the dual structure of an unsupervised paraphrase model to improve semantic parsing.

%% file: conclusion.tex
\section{Conclusion}
\label{sec:conclusion}
In this work, aiming to reduce annotation, we propose a two-stage semantic parsing framework. The first stage utilizes the dual structure of an unsupervised paraphrase model to rewrite the input natural language utterance into canonical utterance. Three self-supervised tasks, namely denoising auto-encoder, back-translation and dual reinforcement learning, are introduced to iteratively improve our model through pre-training and cycle learning phases. Experimental results show that our framework is effective, and compatible with supervised training.

%% file: appendix.tex
\section{Appendices}
\label{sec:appendix}

\subsection{Model Implementations}
\label{app:model}
In this section, we give a full version discussion about all models used in our two-stage semantic parsing framework.
\paragraph{Unsupervised paraphrase model}
We use traditional attention~\citep{luong2015effective} based Seq2Seq model. Different from previous work, we remove the transition function of hidden states between encoder and decoder. The initial hidden states of decoders are initialized to $0$-vectors. Take $D_z\circ E$ paraphrase model as an example: 

(1) a shared encoder encodes the input utterance $x$ into a sequence of contextual representations $\textbf{h}$ through a bi-directional single-layer LSTM~\citep{hochreiter1997long} network~($\psi$ is the embedding function)
\begin{align*}
\overrightarrow{\textbf{h}_i}=&\text{f}_\text{{LSTM}}(\psi(x_i), \overrightarrow{\textbf{h}}_{i-1}), i=1,\cdots,|x|\\
\overleftarrow{\textbf{h}_i}=&\text{f}_\text{{LSTM}}(\psi(x_i), \overleftarrow{\textbf{h}}_{i+1}), i=|x|,\cdots,1\\
\textbf{h}_i=&[\overrightarrow{\textbf{h}}_i;\overleftarrow{\textbf{h}}_i]
\end{align*}

(2) on the decoder side, a traditional LSTM language model at the bottom is used to model dependencies in target utterance $z$~($\phi$ is the embedding function on target side)
\begin{align*}
\textbf{s}_t=&\text{f}_\text{{LSTM}}(\phi(z_{t-1}),\textbf{s}_{t-1})\\
\textbf{s}_0=&0\textrm{-}vector
\end{align*}

(3) output state $s_t$ at each time-step $t$ is then fused with encoded contexts $\textbf{h}$ to obtain the features for final softmax classifier~($\textbf{v}, \mathbf{W_{\ast}}$ and $\mathbf{b_{\ast}}$ are model parameters)
\begin{align*}
u_i^t=&\textbf{v}^T  \text{tanh}(\textbf{W}_h\textbf{h}_i+\textbf{W}_s\textbf{s}_t+\textbf{b}_{a})\\
a_i^t=&\frac{\text{exp}(u_i^t)}{\sum_{j=1}^{|x|}\text{exp}(u_j^t)}\\
\textbf{c}_t=&\sum_{i=1}^{|x|}a_i^t \textbf{h}_i\\
P(z_t|z_{<t},x)=&\text{softmax}(\textbf{W}_o[\textbf{s}_t;\textbf{c}_t]+\textbf{b}_o)
\end{align*}
In both pre-training and cycle learning phases, the unsupervised paraphrase model is trained for $50$ epochs, respectively. To select the best model during unsupervised training, inspired by \citet{lample2017unsupervised}, we use a surrogate criterion since we have no access to labeled data $(x,z)$ even during validation time. For one natural language utterance $x$, we pass it into the model $D_z\circ E$ and obtain a canonical utterance $\hat{z}$ via greedy decoding. Then $\hat{z}$ is forwarded into the dual paraphrase model $D_x\circ E$. By measuring the BLEU score between raw input $x$ and reconstructed utterance $\hat{x}$, we obtain one metric $BLEU(x,\hat{x})$. In the reverse path, we will obtain another metric by calculating the overall accuracy between raw canonical utterance $z$ and its reconstructed version $\hat{z}$ through the naive semantic parser $P_{nsp}$. The overall metric for model selection is~($\lambda$ is a scaling hyper-parameter, set to $4$ in our experiments)
\begin{multline*}\small
\text{Metric}(\mathcal{X}_{dev},\mathcal{Z}_{dev})=\lambda\cdot \mathbb{E}_{x\sim\mathcal{X}_{dev}}[BLEU(x, \hat{x})]\\
+\mathbb{E}_{z\sim\mathcal{Z}_{dev}}[\mathbb{I}\cdot\{P_{nsp}(z)=P_{nsp}(\hat{z})\}]
\end{multline*}

\paragraph{Auxiliary models} The naive semantic parser $P_{nsp}$ is another Seq2Seq model with exactly the same architecture as $D_{z}\circ E$. We do not incorporate copy mechanism cause it has been proven useless on dataset \textsc{Overnight}~\citep{jia-liang-2016-data}. The language models $LM_x$ and $LM_z$ are all single-layer unidirectional LSTM networks. As for style discriminator $P_{dis}$, we use a CNN based sentence classifier~\citep{kim2014convolutional}. We use rectified linear units and filter windows of $3, 4, 5$ with $10, 20, 30$ feature maps respectively. All the auxiliary models are trained with maximum epochs $100$.

For all models discussed above, the embedding dimension is set to $100$, hidden size to $200$, dropout rate between layers to $0.5$. All parameters except embedding layers are initialized by uniformly sampling within the interval $[-0.2,0.2]$.